\documentclass{Interspeech}
\usepackage[utf8]{inputenc}
\usepackage{multirow}
\usepackage{tipa} %
\newcommand\anonymize[1]{[ANONYMIZED]}

\interspeechcameraready

\title{Automatic Speech Recognition of African American English: Lexical and Contextual Effects}

\author[affiliation={1}]{Hamid}{Mojarad}
\author[affiliation={1,2}]{Kevin}{Tang}

\affiliation{Department of English Language and Linguistics}{Institute of English and American Studies, Faculty of Arts and Humanities, Heinrich Heine University Düsseldorf}{Germany}
\affiliation{Department of Linguistics}{University of Florida}{United States of America}
\email{hamid.mojarad@hhu.de, kevin.tang@hhu.de}


\usepackage{comment}

\begin{document}

\maketitle

\begin{abstract}
    
    Automatic Speech Recognition (ASR) models often struggle with the phonetic, phonological, and morphosyntactic features found in African American English (AAE). This study focuses on two key AAE variables: Consonant Cluster Reduction (CCR) and ING-reduction. It examines whether the presence of CCR and ING-reduction increases ASR misrecognition. Subsequently, it investigates whether end-to-end ASR systems without an external Language Model (LM) are more influenced by lexical neighborhood effect and less by contextual predictability compared to systems with an LM. The Corpus of Regional African American Language (CORAAL) was transcribed using wav2vec 2.0 with and without an LM. CCR and ING-reduction were detected using the Montreal Forced Aligner (MFA) with pronunciation expansion. The analysis reveals a small but significant effect of CCR and ING on Word Error Rate (WER) and indicates a stronger presence of lexical neighborhood effect in ASR systems without LMs. 
    
\end{abstract}

\section{Introduction}

Addressing racial bias in ASR has recently become a significant area of concern. Given AAE as a minority dialect, this issue was phonetically confirmed by Koenecke et al. \cite{koenecke2020racial}, who found that the average WER for white American speakers was significantly lower than that for AAE speakers across five prominent ASR systems. Morphosyntactic disparities were further emphasized by Martin \& Tang \cite{martin20_interspeech}, whose examination of “habitual be”, a common AAE morphosyntactic feature, revealed increased WER in ASR performance. Phonological disparities were also underscored by Wassink et al. \cite{wassink2022uneven} in a study of four ethnic dialects from the American Pacific Northwest, which demonstrated higher WER for AAE speakers. 
\subsection{AAE phonological features}

\subsubsection{Consonant cluster reduction}
In the error classification conducted by Wassink et al. \cite{wassink2022uneven}, consonant cluster reduction was identified as the most frequent feature contributing to AAE-related errors. However, the specific origins of these errors remain underexplored. CCR is defined as the simplification of word-final consonant clusters, typically involving the omission of the final stop in a cluster of two consonants (e.g., \textit{cold} /\textipa{koUld}/ → [\textipa{koUl}]) or the penultimate consonant in a cluster of three (e.g., \textit{fists} /\textipa{fIsts}/ → [\textipa{fIs:}]) \cite{Thomas_Bailey}. Given the greater prevalence of two-consonant clusters compared to three-consonant clusters in English \cite{Gregov2010ACA}, this study concentrates specifically on two-consonant clusters ending in a final stop.

\subsubsection{ING-reduction}
In sociolinguistic research, ING-reduction refers to pronunciation variation in words ending with -ing, focusing on the realization of the final nasal segment \cite{Kendall_ing_2021}. This variation manifests in two primary forms: the standard -ing pronunciation with a velar nasal [\textipa{N}] and the reduced -in pronunciation with an alveolar nasal [n]. It occurs both within individual morphemes (e.g., the progressive suffix -ing) and within larger word forms (e.g., something, during), while monosyllabic words (e.g., thing, king) are excluded as they do not exhibit this variability.

\subsection{Lexical and contextual influences}
\subsubsection{Lexical neighborhood effect}
Lexical neighborhood density plays a crucial role in speech recognition, influencing both human perception and ASR systems \cite{Luce1998,jyothi-livescu-2014-revisiting}. This phenomenon has been studied in the context of human speech perception, where words with many lexical neighbors are typically recognized less accurately and more slowly than those with fewer neighbors \cite{Luce1998}. Research on ASR  has shown that lexical neighborhood measures can be predictors of recognition errors, with words having strong competitors in similar contexts being more prone to misrecognition \cite{jyothi-livescu-2014-revisiting, GOLDWATER2010181}.

Phonological reduction, where words are pronounced in a shortened or simplified form, can lead to non-word percepts that are more prone to misrecognition. These reduced forms often have denser lexical neighborhoods due to their shorter length, which increases the challenge of accurate perception \cite{Marian_CLEARPOND}. For instance, the reduced form of ``test'' [\textipa{tEs}] has 21 neighbors, including words like `guess' and `ten', as identified using the CLEARPOND database \cite{Marian_CLEARPOND}. This interaction between phonological reduction and lexical neighborhood density underscores the complex challenges faced by both human listeners and ASR systems in accurately perceiving speech, particularly in casual or fast-paced conversational contexts \cite{Luce1998,GOLDWATER2010181}.

\subsubsection{Contextual predictability}
Contextual predictability is recognized as a critical factor in improving the performance of ASR models by integrating contextual knowledge or text adaptation mechanisms \cite{huang23d_interspeech}. Recent research has shown that incorporating such mechanisms, similar to human cognitive processes, can improve transcription accuracy \cite{Nguyen_Thach}. In ASR systems integrated with language models, the LM plays a role in predicting words based on contextual information, especially when faced with challenges such as degraded acoustic signals, out-of-vocabulary words, or ambiguous phonetic sequences \cite{fox22_interspeech,liu-etal-2024-lost}. In these cases, when the acoustic signal is unclear, like human listeners \cite{Podlubny2018}, LM may prioritize a word that has high contextual predictability given the sentence context. By dynamically adapting to contextual cues, LMs can improve transcription accuracy, particularly for words that are challenging to recognize based solely on acoustic information \cite{Tang_Tung}.  This context-based prediction approach transforms ASR systems from purely acoustic-driven models to more intelligent, context-aware transcription tools that can handle complex linguistic scenarios with greater precision and adaptability. %

\subsection{Present study}
Given that as much as 20\% of ASR errors can be accounted for by sociolinguistic phonological variables \cite{wassink2022uneven}, this study focuses on two common AAE phonological features, namely, CCR and ING-reduction. Building upon Wassink et al. \cite{wassink2022uneven}, we investigate how these variables affect ASR performance using a larger sample of AAE, and whether the resulting errors can be predicted by lexical neighborhood effect and contextual predictability.

In this study, we propose two main hypotheses (H1 and H2). In H1, we hypothesize that the presence of CCR and ING-reduction features in AAE leads to increased ASR misrecognition and higher WER. Building on this, our second hypothesis (H2) presumes that integrating an external LM into state-of-the-art ASR will lead to fewer lexical neighborhood errors. Our error analysis approach, centered on these common phonological features of AAE, aims to quantify the extent to which an external LM can improve ASR performance by reducing lexical neighborhood errors. Data and code are available on \url{osf.io}\footnote{\url{https://doi.org/10.17605/OSF.IO/QN6A2}}.

\section{Methodology}
\subsection{Corpus}
The Corpus of Regional African American Language (CORAAL) \cite{farrington_corpus_2021} serves as the foundational dataset for this study, offering a comprehensive documentation of regional African American Language (AAL) varieties. The corpus provides rich linguistic resources, including audio recordings with time-aligned orthographic transcriptions in TextGrid format, featuring speaker-specific tiers at both utterance and word/phone alignment levels. For this research, we specifically utilized the DCA (Washington, DC) subcorpus, which comprises 74 recordings from 68 speakers (40 men and 28 women) represented in four age groups including ag1 (under 19), ag2 (20-29), ag3 (30-50), and ag4 (51 and over), and also three socioeconomic classes (1 to 3, lowest to highest). The dataset encompasses 34 hours of sociolinguistic interviews, totaling 333,500 words, recorded in WAV (44.1 kHz, 16-bit, mono).

\subsection{Feature extraction}
To extract CCR and ING-reduction features, we employed forced alignment, an approach inspired by Kendall et al.\cite{Kendall_ing_2021}. Their study compared human coding of the sociolinguistic variable (ING) with force alignment and machine learning classifiers, demonstrating that automated coding algorithms can perform close to human coders in their ability to categorize the ING variation. Following this lead, we utilized the Montreal Forced Aligner (MFA, version 2.2.17) \cite{mcauliffe17_interspeech} and the Carnegie Mellon University (CMU) Pronouncing Dictionary to automate the feature extraction process for our analysis.
We identified missing words in our dataset, trained a grapheme-to-phoneme (g2p) model based on the CMU dictionary, and generated pronunciations for these words. We then updated the CMU dictionary with these additions. For words prone to CCR or ING variation, we included both original and reduced pronunciations in the dictionary (e.g., ``accept'' was represented as both ``AH0 K S EH1 P T'' and ``AH0 K S EH1 P''). Using MFA's train command, we developed a custom acoustic model on our entire DCA audio set. Finally, we aligned the complete audio set using this trained acoustic model and the updated CMU dictionary.

\subsection{ASR transcription} 
We employed wav2vec 2.0 \cite{Baevski} as one of the end-to-end ASR models to transcribe our data, specifically using the pretrained model \textit{facebook/wav2vec2-large-960h}. This version, trained on 960 hours of speech, was subsequently enhanced with an external 5-gram LM trained on CORAAL's DCA and DCB subcorpora (entire Washington DC data) using KenLM \cite{heafield-2011-kenlm}. To enable transcription with and without LM, we first resampled our audio files to 16 kHz for compatibility with wav2vec 2.0. The audio was then segmented into chunks of at least 30 seconds, focusing on CCR and ING-reduction prone words, while ensuring no sentences were split mid-utterance by leveraging CORAAL’s provided time frames for speaker utterances.
The segmented audio chunks were processed through wav2vec 2.0 both with and without the LM. Subsequently, we aligned the transcriptions with their corresponding ground truth using the Needleman-Wunsch algorithm for sequence alignment, implemented via the Python \textit{string2string}\footnote{\url{https://github.com/stanfordnlp/string2string}} library. This alignment allowed us to evaluate transcription accuracy.%
\subsection{Post processing}
To ensure a focused analysis of AAE features, we exclusively processed utterances from interviewees, excluding those of interviewers from the DCA dataset. Utilizing the phone alignment obtained in the previous phase, we extracted the corresponding transcription for each target word and its associated utterance. This data was used for calculating WER and examining potential lexical neighborhood effect.

Following Luce's \cite{Luce1986} definition, we considered a word a lexical neighbor if it could be derived from the target word through a single phoneme substitution, deletion, or addition in any position. To identify these neighbors, we employed the Levenshtein algorithm \cite{Levenshtein1966}. Our process involved first checking if the ASR transcribed word existed in the CMU dictionary. For words not found, we generated pronunciations using MFA's g2p command, and then updated the CMU dictionary with the new entries. We then calculated the phonological Levenshtein distance to compile a list of lexical neighbors for each target word. We used the \textit{MFA\_Status} of each target word to determine whether it was detected by MFA in its original or reduced form. If original, we generated the list of neighbors according to the word's full pronunciation. Otherwise (for reduced form detection), we obtained the list of neighbors based on the word's reduced pronunciation. Eventually, if the transcribed word appeared among these neighbors, we attributed the transcription error to the lexical neighborhood effect.

\section{Analyses}
\subsection{H1: phonological reduction increases ASR errors}
\subsubsection{Variables}
In H1, \textit{WER} was analyzed as the dependent variable, with \textit{MFA\_Status}, \textit{AgeGroup}, and \textit{Gender} serving as fixed effect variables. \textit{MFA\_Status} served as a binary factor indicating whether the pronunciation was detected as original or reduced by MFA. 

To address potential non-independence in the data, particularly the influence of frequently occurring target words and individual speaker characteristics, linear mixed-effects regression was employed with \textit{Target\_Word} and \textit{Speaker\_Id} as random effects. Furthermore, for the \textit{Target\_Word} random effect, we included random slopes for \textit{MFA\_Status}, \textit{AgeGroup}, and \textit{Gender}. This means that the impact of pronunciation style, age group, and gender on WER was allowed to vary for different target words. Likewise, for the \textit{Speaker\_Id} random effect, we included a random slope for \textit{MFA\_Status}, which allows the effect of pronunciation style to vary across individual speakers. The analysis was conducted on three datasets (overall, CCR only, and ING only) with two ASR types (with and without LM). The descriptive statistics of the WER is visualized in 
Figure \ref{fig:speech_production}.

\subsubsection{Statistical procedure}
Linear mixed-effects models were fitted using \texttt{lmer} function from the \texttt{lmerTest} package in R (Version 4.4.1) \cite{R_project}. 
The models specify that WER is predicted by the fixed effects of \textit{MFA\_Status}, \textit{AgeGroup}, and \textit{Gender}, with random intercepts and slopes for these predictors across \textit{Target\_Word}, and random intercepts and slopes for \textit{MFA\_Status} across \textit{Speaker\_Id}.\footnote{The model formula:  
\texttt{WER $\sim$ MFA\_Status + AgeGroup + Gender + (1 + MFA\_Status + AgeGroup + Gender | Target\_Word) + (1 + MFA\_Status | Speaker\_Id)}.} The categorical variables were contrast coded.
\textit{MFA\_Status} was sum coded as $-0.5$ for ``Original Pronunciation'' and $0.5$ for ``Reduced Pronunciation''. \textit{Gender} was coded as $-0.5$ for male and $0.5$ for female. \textit{AgeGroup} was Helmert coded to compare each level with the mean of the previous levels.

\subsubsection{Summary of the results}
As shown in Figure \ref{fig:speech_production}, the descriptive statistics of the results suggest a reduced pronunciation leads to a higher WER, but only for CCR, based on the median values.
However, across all datasets - Overall, CCR, and ING in Table 1 - \textit{MFA\_Status} shows a statistically significant positive effect on WER. This effect persists both with and without LM, suggesting that these variations pose consistent challenges for ASR systems. While the impact is statistically significant, the relatively small effect sizes ($\hat{\beta}$ values ranging from 0.021 to 0.040) indicate a moderate rather than severe influence on recognition accuracy. \textit{AgeGroup} appears to have a significant effect on ASR performance, when comparing the second age group to the first (highest $\hat{\beta}$ value in ING dataset without LM). \textit{Gender}, however, does not significantly affect ASR performance on CCR/ING-prone words. %

\begin{table*}[t]
\centering
\caption{Summary of Fixed Effects Across Datasets and ASR Types}
\label{tab:summary}
\resizebox{\textwidth}{!}{%
\begin{tabular}{l cc cc cc}
\toprule
\multirow{2}{*}{\textbf{Effect}} &
\multicolumn{2}{c}{\textbf{Overall Dataset}} &
\multicolumn{2}{c}{\textbf{CCR Dataset}} &
\multicolumn{2}{c}{\textbf{ING Dataset}} \\
\cmidrule(lr){2-3} \cmidrule(lr){4-5} \cmidrule(lr){6-7}
& \textbf{Without LM} & \textbf{With LM} & \textbf{Without LM} & \textbf{With LM} & \textbf{Without LM} & \textbf{With LM} \\
\midrule
MFA\_Status & 0.021 (0.006)$^{***}$ & 0.030 (0.006)$^{***}$ & 0.026 (0.007)$^{***}$ & 0.031 (0.007)$^{***}$ & 0.040 (0.012)$^{**}$ & 0.030 (0.012)$^{*}$ \\
Age Group (2 vs. 1) & -0.158 (0.042)$^{***}$ & -0.108 (0.035)$^{**}$ & -0.146 (0.042)$^{***}$ & -0.102 (0.034)$^{**}$ & -0.173 (0.046)$^{***}$ & -0.131 (0.039)$^{**}$ \\
Age Group (3 vs. 2,1) & -0.041 (0.030) & -0.035 (0.025) & -0.038 (0.030) & -0.032 (0.024) & -0.035 (0.033) & -0.029 (0.027) \\
Age Group (4 vs. 3, 2, 1) & 0.033 (0.029) & 0.037 (0.024) & 0.038 (0.028) & 0.033 (0.023) & 0.030 (0.031) & 0.035 (0.026) \\
Gender (Female vs. Male) & -0.013 (0.033) & -0.028 (0.027) & -0.011 (0.033) & -0.029 (0.027) & -0.003 (0.037) & -0.013 (0.031)\\
\bottomrule
\end{tabular}%
}
\parbox{\textwidth}{\small Note: Values are presented as: Estimate (Standard Error). 
Significance levels: *** p \textless{} 0.001, ** p \textless{} 0.01, * p \textless{} 0.05.}
\end{table*}

\subsection{H2: LM reduces ASR neighborhood errors}
\subsubsection{Variables}
In the second hypothesis, \textit{Neighborhood\_Status} was analyzed as the dependent variable, with  \textit{ASR\_Type} being the fixed effect variable. \textit{Neighborhood\_Status} was coded as binary variable (reference level: \textit{Neighbor\_Error}), and contrast coding was applied to \textit{ASR\_Type} (\textit{without\_LM}: -0.5, \textit{with\_LM}: 0.5).
Mixed-effects logistic regression was employed with \textit{Target\_Word} and \textit{Speaker\_Id} as random effects. Furthermore, for both random effects, we included random slopes for \textit{ASR\_Type} to allow for the impact of ASR type on \textit{Neighborhood\_Status} to vary for different target words and across individual speakers.
\subsubsection{Statistical procedure}
A Logistic mixed-effects model was fitted using the \texttt{glmer} function from the \texttt{lme4} package in R.
A logit link function was chosen since the \textit{Neighborhood\_Status} variable is binary (\textit{Non\_Neighbor\_Error} vs. \textit{Neighbor\_Error}). The model was applied to the merged dataset combining both ASR types. To implement it, we filtered out the correctly transcribed ASR words for our target words to obtain only the errors.\footnote{The model formula:
\texttt{Neighborhood\_Status $\sim$ ASR\_Type + (1 + ASR\_Type | Target\_Word) + (1 + ASR\_Type | Speaker\_Id).}}

\subsubsection{Summary of the results}
In ASR without an LM, we observed 7.9\% (1,006) of neighborhood errors out of 12,734 total incorrect transcriptions for our target words. However, with the integration of an LM, the number of neighborhood errors drastically decreased to 3.3\% (277) out of the total misrecognitions of 8,283. This descriptive finding is confirmed by the regression model, which reveals significant effects across all datasets, indicating that language model usage influences lexical neighborhood errors. \textit{ASR\_Type} shows a consistent, significant positive effect ($p$s $< 0.001$) when comparing ASR with and without LM. This effect is strongest for the ING dataset $(\hat{\beta}: -2.1879)$, followed by the overall dataset $(\hat{\beta}: -1.2875)$, and the CCR dataset $(\hat{\beta}: -0.8954)$.

\begin{figure}[t]
  \centering
  \includegraphics[width=\linewidth]{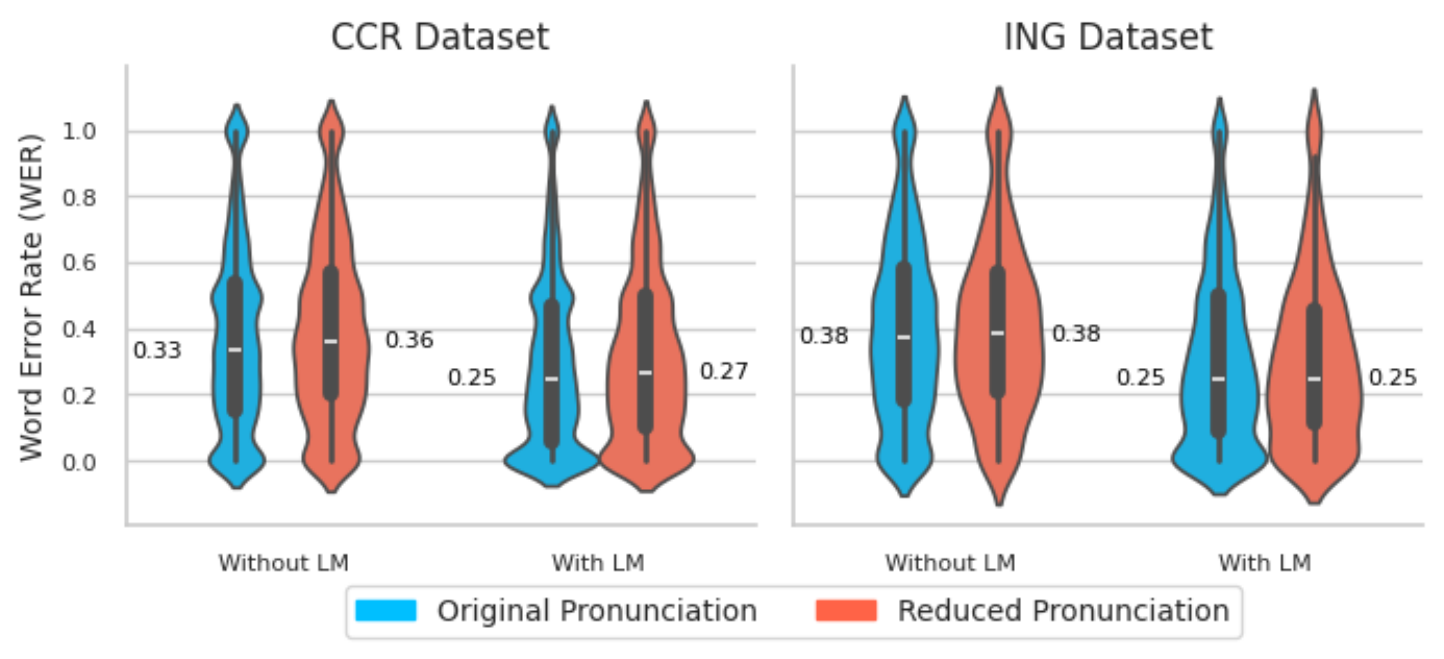}
  \caption{WER by MFA Status for CCR and ING Target Words}
  \label{fig:speech_production}
\end{figure}

\section{Discussion}
Our study reveals notable insights into the performance of ASR systems when confronted with CCR and ING-reduction, as common AAE variations. The consistent positive effect of \textit{MFA\_Status} across datasets indicates that AAE features significantly influence ASR misrecognition. This effect still remains significant even when we recruit an external LM to provide further context for ASR to generate more accurate transcriptions. Therefore, this strongly supports our first hypothesis, which proposed that the presence of CCR and ING-reduction variations contributes to increased ASR misrecognition.

Expanding on this finding, our WERs detailed in Figure \ref{fig:speech_production} are comparable to previous reports on the wav2vec 2.0 model on the AAE datasets. Johnson et al. \cite{Johnson2023Analysis}, for instance, reported WERs of 39\% for story retelling and 30\% for picture description tasks when using wav2vec 2.0 on AAE speech. Similarly, Chang et al. \cite{chang24d_interspeech} found a 52.8\% WER for wav2vec 2.0 transcriptions of the CORAAL dataset, and emphasized that utterances with more phonological and morphosyntactic AAE features exhibited higher error rates. These findings align with our results and highlight the challenges in recognizing AAE speech, and underscore the need for further model adaptation to improve dialectal diversity handling.

The observed age-related effects on ASR performance highlight generational variations in language use within AAE-speaking communities. Younger speakers exhibited higher WER than older speakers across datasets, suggesting that CCR and ING-reduction variations pose additional challenges for current ASR systems. This finding contrasts with the broader understanding of ASR performance, which typically shows higher WER for children \cite{GurunathShivakumar} and elderly speakers \cite{KWON2016110} due to factors such as articulatory variability and slower speaking rates. In our case, the \textit{ag1} group is more accurately described as adolescents or teenagers rather than children, as the speakers were recorded between 1968 and 1969, with birth dates ranging from 1891 to 1958 \cite{farrington_corpus_2021}. This means that the youngest speaker would have been at least 10 years old at the time of recording.

In contrast to several previous studies that have reported gender-based disparities in ASR performance, our research found no significant effect of gender on recognition accuracy. This finding diverges from the existing literature, which has often shown mixed results with some studies favoring male speakers \cite{tatman17_interspeech} and others indicating better performance for female speakers \cite{koenecke2020racial, harris-etal-2024-modeling}. This finding suggests that gender-based variability may not play a substantial role in ASR performance for AAE speakers, at least within the scope of this study.

In our second hypothesis, we argued that integrating an external LM into the ASR model would reduce errors stemming from lexical neighborhood effect. This was strongly supported by our findings in Section 3.2.3. In other words, while end-to-end ASR models are often promoted for their ability to eliminate the need for separate LMs \cite{Graves2014}, our results align with recent research \cite{Nguyen_Thach, huang23d_interspeech, liu-etal-2024-lost} that underscores the continued importance of LMs in improving ASR performance. As also illustrated in Figure \ref{fig:speech_production}, incorporating an LM significantly reduced WER for both the CCR and ING datasets. This reduction can be attributed to the LM's ability to provide contextual predictability, thereby mitigating the lexical neighborhood effect.

Additionally, the study revealed that non-neighbor errors were considerably more frequent than neighbor errors, particularly in ASR systems without LMs. This suggests that there are still other factors that could be driving the errors, such as the general limited amount of training data, the mismatches in the acoustics of the training data and the test data \cite{koenecke2020racial}, and other dialectal features that we have not considered \cite{martin20_interspeech}.

One key implication of our findings is that annotating phonological variations during training could enhance ASR accuracy by explicitly capturing the acoustic variability in AAE. For example, some efforts have been made in automatic feature annotation of AAE (see \cite{porwal-etal-2025-analysis} and references therein). Such annotations would help ASR systems better account for systematic phonological differences like CCR and ING-reduction, thereby improving accuracy and reducing bias against underrepresented speech communities.

Several limitations of our study can be addressed in the future. Firstly, due to time constraints, we were unable to evaluate MFA detection of CCR and ING-reduction variables with human coding. This comparison, as done by Kendall et al. \cite{Kendall_ing_2021} for ING-reduction in CORAAL, could have enhanced the generalizability of our CCR results. Secondly, we used the Large-960h wav2vec 2.0 model, due to its compatibility with external language models, to address the second hypothesis; however, evaluating models with more training hours could lower error rates, and provide a clearer picture of the lexical neighborhood errors. This requirement also limited our model choices for testing the first hypothesis, as including additional ASR models would improve the generalizability of the study. Finally, we did not explicitly test whether the effect of CCR/ING-reduction on ASR performance is influenced by an increase in lexical neighborhood density. Instead, we relied on the well-established relationship between word length and neighborhood size.

\section{Conclusion}
This study examined the performance of ASR systems, focusing on CCR and ING-reduction, two common phonological variations in AAE. Our findings underscore the persistent challenges ASR systems face when transcribing dialectal speech, even with advanced architectures like wav2vec 2.0 and the integration of LMs. First, our results confirmed that CCR and ING-reduction variations significantly contribute to ASR misrecognitions, strongly supporting our initial hypothesis. To further explore this, we analyzed the effects of gender and age on ASR performance. While gender showed no significant impact, age was a critical factor among AAE speakers under 19, leading to higher rates of ASR misrecognition. Second, across all datasets, ASR with an LM consistently outperformed the one without an LM in reducing neighborhood errors. This validates our second hypothesis and highlights the LM's ability to leverage contextual predictability, minimize confusion between phonetically similar words, and improve transcription accuracy.

\section[Acknowledgements]{Acknowledgements\protect\footnote{\url{https://credit.niso.org/}}}

The work is part of HM's PhD. Conceptualization, Methodology, Formal Analysis, Writing – original draft/review/editing: HM, KT; Data Curation, Investigation, Funding Acquisition, Project Administration, Software, Validation: HM; Resources, Supervision: KT.

\bibliographystyle{IEEEtran}
\bibliography{mybib}

\end{document}